\documentclass[runningheads]{llncs}
\usepackage{graphicx}
\usepackage{amsmath,amssymb} 
\usepackage{color}
\usepackage[width=122mm,left=12mm,paperwidth=146mm,height=193mm,top=12mm,paperheight=217mm]{geometry}
\usepackage{epsfig}
\usepackage{lipsum}
\usepackage[perpage,symbol]{footmisc}
\usepackage{mathrsfs}
\usepackage{multirow}
\usepackage{array}
\newcommand{\bd}[1]{\textbf{#1}}
\newcommand{\etal}{\emph{et al.}}
\newcommand{\eg}{\emph{e.g. }}
\newcommand{\ie}{\emph{i.e. }}
\newcommand{\vs}{\emph{vs. }}
\newcommand{\tablestyle}[2]{\setlength{\tabcolsep}{#1}\renewcommand{\arraystretch}{#2}\centering\footnotesize}
\usepackage[pagebackref=true,breaklinks=true,letterpaper=true,colorlinks,bookmarks=false]{hyperref}
\begin{document}
\pagestyle{headings}
\mainmatter

\title{Quantization Mimic: Towards Very Tiny CNN for Object Detection} 

\titlerunning{Quantization Mimic}

\authorrunning{Wei \etal}
\author{Yi Wei$^1$$^\dagger$, Xinyu Pan$^2$$^\dagger$, Hongwei Qin$^3$, Wanli Ouyang$^4$, Junjie Yan$^3$}
\footnotetext[2]{The work was done during an internship at SenseTime}


\institute{$^1$Tsinghua University, Beijing, China\\$^2$The Chinese University of Hong Kong, Hong Kong, China\\$^3$SenseTime, Beijing, China\\$^4$The University of Sydney, SenseTime Computer Vision Research Group, Sydney, New South Wales, Australia\\\texttt{wei-y15@mails.tsinghua.edu.cn,THUSEpxy@gmail.com\\qinhongwei@sensetime.com,wanli.ouyang@sydney.edu.au\\yanjunjie@sensetime.com}}

\maketitle
\renewcommand{\thefootnote}{\arabic{footnote}}
\begin{abstract}
	In this paper, we propose a simple and general framework for training very tiny CNNs (\eg VGG with the number of channels reduced to $\frac{1}{32}$) for object detection. Due to limited representation ability, it is challenging to train very tiny networks for complicated tasks like detection. To the best of our knowledge, our method, called Quantization Mimic, is the first one focusing on very tiny networks. We utilize two types of acceleration methods: mimic and quantization. Mimic improves the performance of a student network by transfering knowledge from a teacher network.  Quantization converts a full-precision network to a quantized one without large degradation of performance. If the teacher network is quantized, the search scope of the student network will be smaller. Using this feature of the quantization, we propose Quantization Mimic. It first quantizes the large network, then mimic a quantized small network. The quantization operation can help student network to better match the feature maps from teacher network. To evaluate our approach, we carry out experiments on various popular CNNs including VGG and Resnet, as well as different detection frameworks including Faster R-CNN and R-FCN. Experiments on Pascal VOC and WIDER FACE verify that our Quantization Mimic algorithm can be applied on various settings and outperforms state-of-the-art model acceleration methods given limited computing resouces. 

\end{abstract}

\section{Introduction}
In recent years, CNN achieved great success on various computer vision tasks. However, due to their huge model size and computation complexity, many CNN models cannot be applied on real world devices directly. Many previous works focus on how to accelerate CNNs. They can be roughly divided to four categories: quantization (\eg BinaryNet \cite{courbariaux2016binarized}), group convolution based method (\eg MobileNet \cite{howard2017mobilenets}), pruning (\eg channel pruning\cite{he2017channel}) and mimic (\eg Li
\etal \cite{li2017mimicking}).


Although most of these works can accelerate models without degradation of performance, their speed-up ratios are limited (\eg compress VGG to VGG-1-4\footnote {In this paper \emph{-1-n} network means a network whose channel numbers of every layer is reduced to $\frac{1}{n}$ compared with original network.}). Few methods are experimented on very tiny models (\eg compress VGG to VGG-1-16). "Very tiny" is a relative concept and we define it as a model whose channel numbers of every layer is less than or equal to $\frac{1}{16}$ compared with original model. Our experiments show that our method outperform other approaches for very tiny models. 

\begin{figure}[tb]
	\centering  
	\includegraphics[ width=0.7\linewidth]{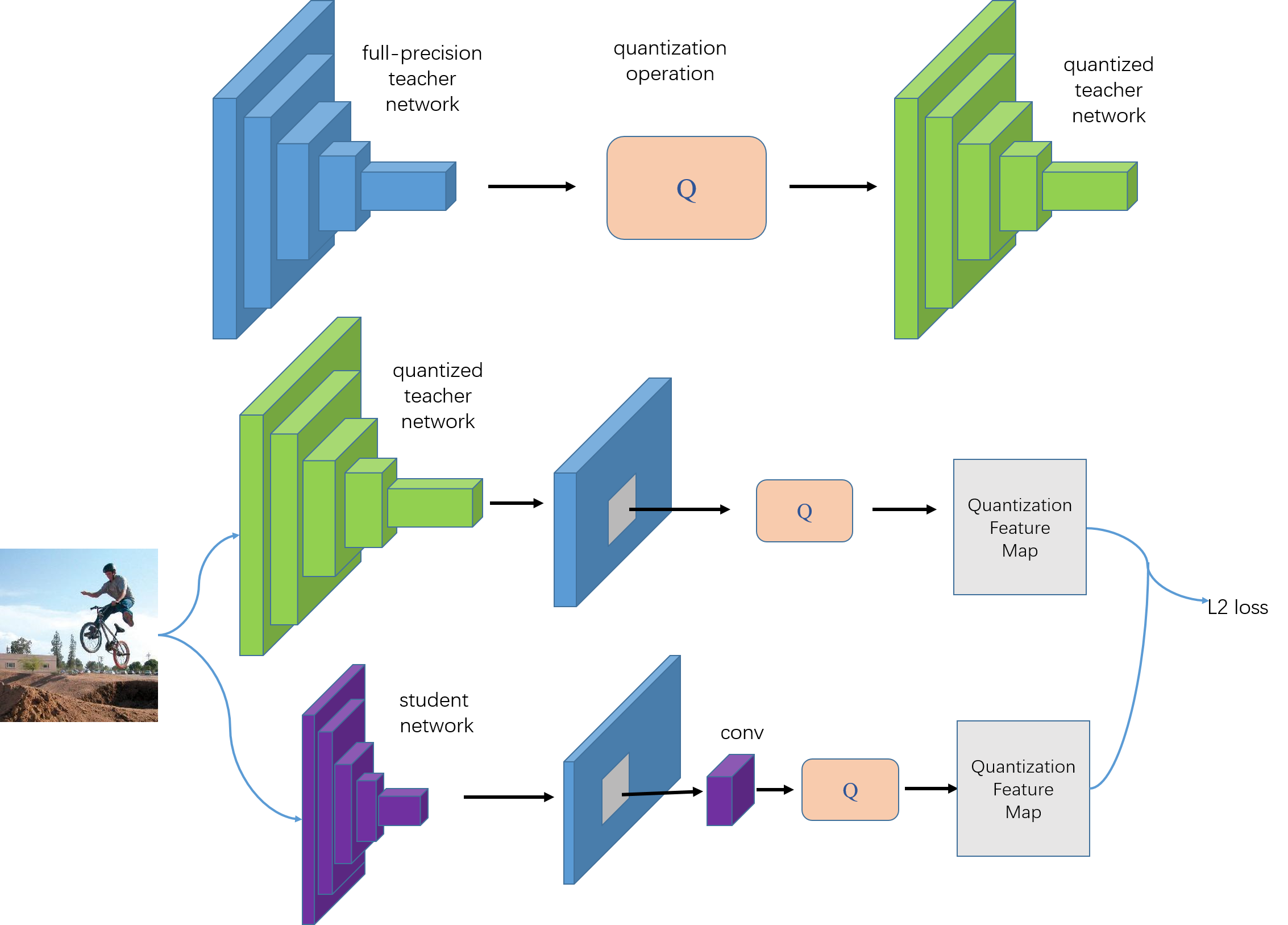}  
	\caption{The pipeline of our method. First we train a full-precision teacher network. Then we operate quantization on the feature map of full-precision teacher network and we get a quantized network. Finally we use this quantized network as teacher model to teach a quantized student network. We emphasize that we do both quantization operation on feature maps of student and teacher networks in training stages. }
	\label{fig:pipeline}  
\end{figure}  

As two kinds of model acceleration methods, quantization and mimic are widely used to compress model. Quantization methods can transfer a full-precision model to a quantized model\footnote{The quantized network in this paper means a network whose output feature map is quantized but not means parameter is quantized} while maintaining similar accuracy. However, using quantization method to directly speed up models usually need extra specific implementation (\eg FPGA) and specific instruction set. Mimic methods can be used on different frameworks and easy to implement. The essence of these methods is knowledge transfer in which student networks learn the high-level representations from teacher networks. However, when applied on very tiny networks, mimic method does not work well either. This is also caused by the very limited representation capacity.

It is a natural hypothesis that if we use quantization method to discretize the feature map of the teacher model, the search scope of the student network will get shrinked and it will be easier to transfer knowledge. And quantization on student network can increase the matching ratio on the discrete feature map from teacher network. In this paper, we propose a new approach utilizing the advantages of quantization and mimic methods to train very tiny networks. Figure \ref{fig:pipeline} illustrates the pipeline.  Quantization operation is applied to the feature map of the teacher model and the student model. The quantized feature map of the teacher model is used as supervision of the student model. We propose that this quantization operation can facilitate feature map matching between two networks and make knowledge transfer easier.

To summarize, the contributions of this paper are as follows:
\begin{itemize}
	\item We propose an effective algorithm to train very tiny networks. To the best of our knowledge, this is the first work focusing on very tiny networks.
	\item We utilize quantized feature maps to facilitate knowledge distilling, \ie quantization and mimic. 
	\item We use a complicated task object detection instead of image classification to verify our method. Sufficient experiments on various CNNs, frameworks and datasets validate our approach effective. 
	\item The method is easy to implement and has no special limitation during training and inference.
\end{itemize}

\section{Related Work}
\subsection{Object Detections}  
The target of object detection \cite{zeng2017crafting,liu2017recurrent,yan2015object,yan2014fastest,ouyang2015deepid,ouyang2017chained} is to locate and classify the objects in images. Before the success of convolutional neural network, some traditional pattern recognition algorithms (HOG \cite{wang2009hog}, DPM \cite{lowe2004distinctive} \etal) are used on this task. Recently, R-CNN \cite{girshick2014rich} and its variants become the popular method for object detection task. The SPP-Net \cite{he2014spatial} and Fast R-CNN  \cite{girshick2015fast} reuse feature maps to speed up R-CNN framework. Beyond the pipeline of Fast R-CNN, Faster R-CNN add region proposal networks and use joint-train method during training. R-FCN utilize position-sensitive score maps to reduce more computation. YOLO \cite{redmon2016you} and SSD \cite{liu2016ssd} are the typical algorithms of region-free methods. Although the frameworks used in this paper are from region proposal solutions family, Quantization Mimic can easily transform to YOLO and SSD methods. 

\subsection{Model Compression and Acceleration}

\subsubsection{Group Convolution Based Methods:} The main point of this kind of methods is to use group convolution for acceleration. Mobilenet \cite{howard2017mobilenets} and Googlenet Xception \cite{chollet2016xception} utilize Depthwise Convolution to extract features and Pointwise Convolution to merge features. Beyond these works, Zhang \etal \cite{zhang2017interleaved} propose a general group convolution algorithm and show that Xception is the special case of their method. Group operation will block the information flow between different group convolutions and most recently, Shufflenet \cite{zhang2017shufflenet} introduces channel shuffle approach to solve this problem.  

\subsubsection{Quantization:} Quantization methods \cite{rastegari2016xnor,zhou2016dorefa} can reduce the size of models efficiently and speed up for special implementation. BinaryConnect  \cite{courbariaux2015binaryconnect}, binarized neural network (BNN) \cite{courbariaux2016binarized} and LBCNN \cite{juefei2016local}  replace floating convolutional filter with binary filter. Furthermore, INQ \cite{zhou2017incremental} introduce a training method to quantize model whose weights are constrained to be either powers of two or zero without a decrease on performance. Despite these advantages, quantization models can only be used to speed up on special devices.

\subsubsection{Pruning and Sparse connection:} \cite{alvarez2016learning,wen2016learning} set sparse constraint during training for pruning. \cite{anwar2016compact,li2016pruning} focus on the importance of different filter weights and do pruning operation according to weights’ importance. And these methods are training-based, which are more costly. Recently He \etal \cite{he2017channel} propose an inference-time pruning method, using LASSO regression and least square construction to select channels in classification and detection task. Furthermore, Molchanov \etal \cite{molchanov2016pruning} combine transfer learning and greedy criteria-based pruning. We use He \etal \cite{he2017channel} and Molchanov \etal \cite{molchanov2016pruning} for comparing our alogrithm and we will show that it is difficult for them to prune a large network (such as VGG) to a very tiny network (such as VGG-1-32). Sparse connection \cite{guo2016dynamic,han2016eie,han2015learning,yang2016designing} can be considered as parameter-wise pruning method, eliminating connection between neurons.

\subsubsection{Mimic:}  The principle of mimic is Knowledge Transfer. As a pioneering work, Knowledge Distillation (KD)  \cite{hinton2015distilling} defines soft targets as outputs of the teacher network. Compared with labels, soft targets provide extra information about inter-class similarities. FitNet \cite{romero2014fitnets} develops  Knowledge Transfer as whole feature map mimic learning to compress wide and shallow networks to thin and deep networks.  Li \etal \cite{li2017mimicking} extend mimic techniques for object detection task. We use their joint-train version as our baseline.


\section{Our Approach}

In this section, we first introduce the quantization method and mimic method we use separately, then combine them and propose the pipeline of Quanzition Mimic algorithm. In \S\ref{sec:analysis} we show the theoretical analysis of our approach. 

\subsection{Quantization} 
\cite{courbariaux2015binaryconnect,rastegari2016xnor,zhou2016dorefa} use quantization method to compress models directly. Unlike them, we use quantization to limit the range and help mimic learning. In details, the quantization for teacher network is to discretize its output and in the meanwhile we can guarantee the accuracy of teacher network when doing quantization. And quantizing the output of student network can help it match the discrete output of teacher network, which is the goal of mimic learning. In our work, we do quantization operation on the last activation layer of the teacher network.

\begin{figure}[tb]
	\centering
	\setlength{\belowcaptionskip}{-5pt}
	\includegraphics[  width=0.5\linewidth]{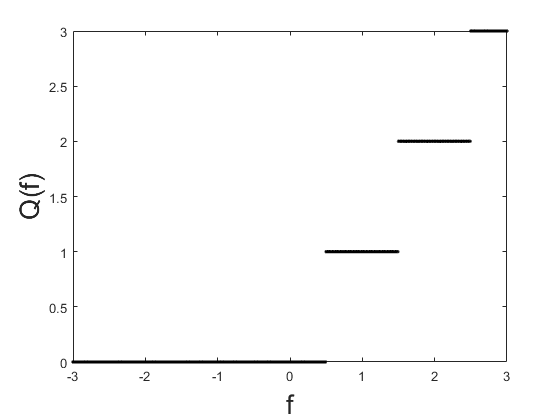}
	\caption{Quantized ReLU function. The new activation function is defined as $\widetilde{f}=Q\left(f\right)$,where $f$ is the original activation function.}
	\label{fig:relu}
\end{figure}

INQ \cite{zhou2017incremental} constrains the output to be either zero or power of two. Different from them, we use uniform quantization for the following reason. R-FCN \cite{dai2016r} and Faster R-CNN \cite{ren2015faster} use RoI pooling operation which is a kind of max pooling operation. The output of RoI pooling  layer is determined by the max response of every block in RoIs. So it is important to describe strong response of feature maps more accurately. Uniform quantization can better describe large value than power of two quantization.
We define the element-wise quantization function $Q$ as:
\begin{equation}
Q\left(f\right)= \beta \quad \text{if} \ {\frac{\alpha+\beta}{2}}<f\leq {\frac{\gamma+\beta}{2}}
\label{eq:Q_function} 
\end{equation}
where $\alpha$ ,$\beta$ and $\gamma$ are the adjacent entries in the code dictionary $D$:
\begin{equation}
D=\left\{0,s,2s,3s….. \right\}
\end{equation}
where s is the stride of uniform quantization.
We use function $Q$ to convert full-precision feature maps to quantized feature maps:
\begin{equation}
\widetilde{f}=Q\left(f\right)
\end{equation}
where $f$ is the feature map. Figure \ref{fig:relu} illustrates quantized ReLU function.
As for backward propagation, inspired by BNN \cite{courbariaux2016binarized}, we use the full-precision gradient. We find that quantized gradient will cause the student network difficult to converge.

\subsection{Mimic} \label{sec:mimic}
In popular CNN detectors, the feature map from feature extractors (\eg VGG, Resnet) will affect both localization and classification accuracy. We use L2 regression to let student networks learn the feature map from the teacher networks and utilize Li \etal \cite{li2017mimicking} joint-train version as our backbone. Unlike soft target \cite{hinton2015distilling} whose dimension is equal to the number of categories, the dimension of feature maps is related to the size of inputs and networks architecture. Sometimes number can be millions. Simply mimicking the whole feature maps is difficult for student network to converge.  Faster R-CNN \cite{ren2015faster} and R-FCN \cite{dai2016r} are region-based detectors and both of them use RoI-Pooling operation. So the region of interest plays more important role than other regions. We use mimic learning between the region of interest on student’s and  teacher’s feature maps. The whole loss function of mimic learning is described as follows. 
\begin{equation}
L=L_{cls}^{r}+L_{reg}^{r}+L_{cls}^{d}+L_{reg}^{d}+\lambda L_{m}
\end{equation}
\begin{equation}
L_{m}={\frac{1}{2N}}\sum_{i}\left\| f_{t}^{i}-r\left(f_{s}^{i}\right) \right\|_{2}^{2}
\end{equation}
where $ L_{cls}^{r}$,$ L_{reg}^{r}$ are the loss function of region proposal networks \cite{girshick2015fast} while $ L_{cls}^{d}$,$ L_{reg}^{d}$ are the function of R-FCN or Faster R-CNN detectors. We define $L_{m}$ as the mimic-loss and $\lambda$ is the loss weight. N is the number of region proposals. $ f_{t}^{i}$ and $ f_{s}^{i}$ represent the $i$th region proposal on teacher and student network’s feature maps. Function $r$ transfers the feature map from student network to the same size of teacher network. The mimic learning is on the last year of feature extractor networks. 

Though RoI mimic learning reduces the dimension of feature maps and helps student network convergence, very tiny network is sensitive to mimic loss weight $\lambda$. If $\lambda$ is small, it will weaken the effectiveness of mimic learning. In the contrast, large $\lambda$ will also bring bad results. Due to the poor learning capacity of very tiny network, large $\lambda$ will cause it focus on the learning of teacher network's feature map at the begining of training. In this way, it will ignore other loss. We name this phenomenon as `gradient focus' and we set $\lambda$ as 0.1, 1 and 10 for experiments.  

\subsection{Quantization Mimic}
\begin{figure}[tb]
	\centering
	\includegraphics[width=0.6\linewidth ]{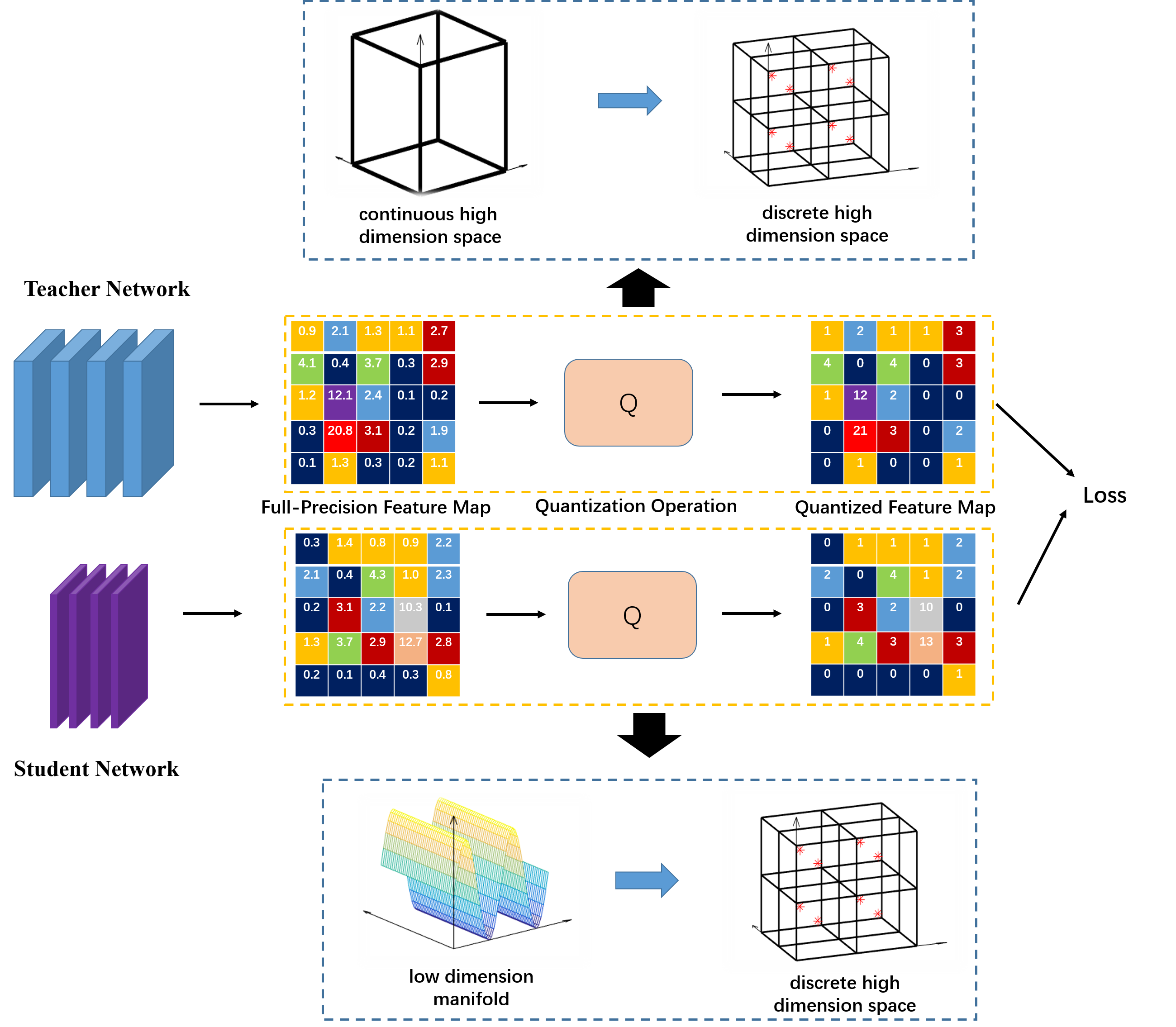}
	\caption{The effect of quantizatuon operation. We use quantized teacher network to guide quantized student network. The quantization on teacher network can discretize its feature maps and convert a continous high dimension space to a discrete high dimension space. And for student network, quantization helps low dimension manifold to match a discrete high dimension feature map. In this way, mimic learning becomes easier .}
	\label{fig:introduction}
\end{figure}

The pipeline of our algorithm is as follows: First we train a full-precision teacher network. Then we use function $Q$ to compress full-precision teacher network to a quantized network. To get high performance compressed model, we finetune on full-precision network. Finally, we utilize quantized teacher network to teach student network using mimic loss as supervision. And during training, we both quantize the feature map of teacher and student network. Figure \ref{fig:introduction} illustrates our method.

Because of quantization operation, the mimic loss $L_{m}$ is redefined as:
\begin{small}
	\begin{equation}
	L_{m}={\frac{1}{2N}}\sum_{i}\left\| Q\left(f_{t}^{i}\right)-Q\left(r\left(f_{s}^{i}\right)\right) \right\|_{2}^{2}
	\end{equation}
\end{small}
where quantization function $Q$ is defined in Equation \ref{eq:Q_function}

\subsection{Analysis} \label{sec:analysis}
We will show that the quantization of both teacher and student networks will facilitate feature maps matching between student and teacher networks and help student network learn better. Figure \ref{fig:introduction} shows the effect of quantization operation. We assume that $f_{t}^{n}$ is the feature map of full-precision teacher network with the input $I_{n}$. The width, height and channel numbers of $f_{t}^{n}$ are $W_{t}^{n}$,$H_{t}^{n}$ and $C_{t}^{n}$.We squeeze $f_{t}^{n}$ as a column vector $y_{n}$ whose dimension is $ W_{t}^{n}H_{t}^{n}C_{t}^{n}$. The target of mimic learning is to get approximate solution of the following equation:
\begin{equation}
Y=w_{s}I
\end{equation}
\begin{equation}
Y=\left[y_{1},y_{2},...,y_{n}\right]
\end{equation}
\begin{equation}
I=\left[I_{1},I_{2},...,I_{n}\right]
\end{equation}
where $w_{s}$ is the weights of student network. However, due to the high dimensionality of $y_{n}$ and large image numbers, the rank of $Y$ can be very high. On the other hand, very tiny networks have few parameters and the rank of $w_{s}$ is low. Therefore, it is difficult for very tiny student networks to mimic high dimension feature maps directly. 
The target of Quantization Mimic is changed as:
\begin{equation}
Q\left(Y\right)=Q\left(w_{s}I\right)
\end{equation}
where $Q$ is quantization function.
The quantization operation on the output of teacher network discretizes its feature maps. Furthermore, because of the range of element in feature maps is bounded, the value of every entry in matrix $ Q\left(Y\right)$ is discrete and finite. For example, if the range of element in $f_{t}^{n}$ is $\left[0,40\right]$ and the stride of uniform quantization is 8, the possible value of entry in $ Q\left(Y\right)$ is from $\left\{0,8,16,24,32,40\right\}$. In this way, we convert continuous high dimension space to discrete high dimension space. 

The quantization on student networks makes it easier to match the $Q\left(f_{t}^{n}\right)$. Every axis of target space for student network can be separated by entries in code dictionary. And the whole space is separated by several high dimension cubes. 

\begin{figure}[tb]
	\setlength{\belowcaptionskip}{-5pt}
	\centering  
	\includegraphics[ width=0.46\linewidth]{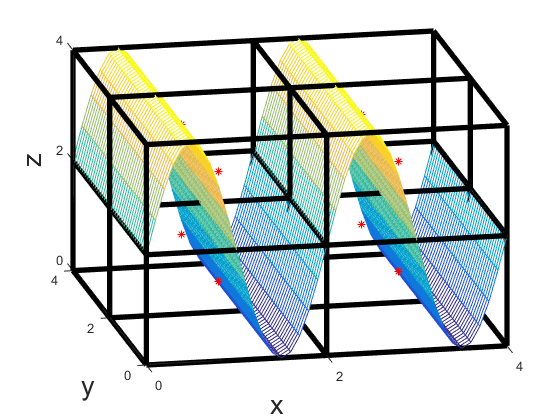}  
	\caption{A manifold in 3-dimension space. The manifold intersect all 8 cubes. The point '*' represent the center of cube, which is the vector after quantization operation. }  
	\label{fig:cube} 
\end{figure}

For simplicity, we assume the dimension of target space $\phi$ is 3, \ie, the dimension of $y_{n}$ is 3. The code dictionary is selected as $\left\{1,3\right\}$. Because of quantization operation, this 3-dimension space is separated by 8 cubes (See Figure~ \ref{fig:cube}). If a vector $v$ is in cube $c$ , after quantization operation, it will be the center of cube $c$. For example, $v=\left[1.2,2.2,1.8\right]^\mathrm{T}$,$Q\left(v\right)=\left[1,3,1\right]^ \mathrm{T}$, and $\left[1,3,1\right]^ \mathrm{T}$ is the center of a cube. 

We suppose that feature maps of student network consist a low dimension manifold. The goal of mimic learning is to use this manifold to fit all 8 cube centers, \ie, we want these 8 centers on the manifold. However,  after introducing quantization on student network, if the manifold intersect a cube, the manifold can achieve the center of this cube. Thus, instead of matching all centers, we only need the manifold to intersect 8 cubes, which weaken matching conditions. And in this way, there are more suitable manifolds , which promotes feature maps matching between two networks. Experiments in \S\ref{ablation_quantization} shows that our approach is still effective in high dimension case. Figure \ref{fig:cube} illustrates a manifold in 3-dimension space which intersect all cubes.

\subsection{Implementation Details}
We train networks with Caffe \cite{jia2014caffe} using C++ on 8 Nvidia GPU Titan X Pascal. We use stochastic gradient descent (SGD) algorithm. The weight decay is 0.0005 and momentum is 0.9. We set uniform quantization stride as 1 for all experiments.

\paragraph{VGG with R-FCN:} In this experiment we rescale the images such that their shorter side is 600 and we use original images for test. We use gray images as input. The learning rate is 0.001 for the first 50K iterations and 0.0001 for the next 30K iterations. The teacher network is VGG-1-4 with R-FCN and we set mimic loss weight $\lambda$ as 1. For RPN anchors, we use one aspect ratio and 4 scales with box areas of $4^{2}$, $8^{2}$, $16^{2}$, $32^{2}$. 2000 RoIs are used to sample the features on the feature maps of teacher and student networks. The ROI output size of R-FCN detector is set as $3\times3$. We utilize OHEM \cite{shrivastava2016training} algorithm to help training. 

\paragraph{Resnet with Faster R-CNN:} We rescale all the images such that shorter side is 600 for both training and test. We totally train 40K iterations. The learning rate is 0.001 for the first 30K iterations and 0.001 for the last 10k iterations. We set $\lambda$ as 0.1, 1 and 10 for Resnet experiment respectively. And for RPN anchors, we use 2 aspect ratios (2:1, 3:1) and 3 scales with box areas of $4^{2}$,  $8^{2}$ and $16^{2}$. 128 RoIs are used to sample the features on the feature maps of teacher and student networks. The ROI output size of Faster R-CNN detector is set as $7\times7$.

\section{Experiments}
To prove the generalization ability of our method, we evaluate our approach for different frameworks on different datasets. In detail, we use VGG with R-FCN and Resnet with Faster R-CNN as our backbones. Results are reported on WIDER FACE \cite{yang2016wider} and Pascal VOC \cite{everingham2010pascal}.

\subsection{Experiments on WIDER FACE Dataset}
WIDER FACE dataset \cite{yang2016wider} contains about 32K images with 394K annotated faces. The size of faces in WIDER FACE dataset vary a lot. The validation and Test set are divided into \emph{easy} , \emph{medium} and \emph{hard} subsets. 
We find that VGG and VGG-1-4 have similar performance on WIDER FACE dataset (See Table \ref{tab:vgg_self}) and we use VGG-1-4 with R-FCN detector as our teacher network  (large model). To show the superiority of our algorithm, VGG-1-32 with R-FCN detector is selected as the student network (small model). Table \ref{tab:speed_vgg} illustrate the speed and size of our very tiny model student network compared with large models. It has extremely small size and fast speed.
\begin{table}[h]
	\newcommand{\tabincell}[2]{\begin{tabular}{@{}#1@{}}#2\end{tabular}}
	\centering
	\caption{ The comparision between VGG, VGG-1-4, VGG-1-32 with R-FCN detector on speed and size. The size is calculated theoretically. VGG-1-32 with R-FCN has tiny size and amazing speed, which can be applied on embedded devices. Tested on Titan X GPU with a single image of which the longer side is resized to 1024.}
	\begin{tabular}{c|c|c}
		\hline
		Method & Speed & Size \\
		\hline \hline
		\tabincell{c}{VGG  \\ with R-FCN} & 103.6ms & 79.8M \\
		\hline
		\tabincell{c}{VGG-1-4\\ with R-FCN} & 30.2ms & 5.04M \\
		\hline
		\tabincell{c}{VGG-1-32 \\with R-FCN} & \bd{9.6ms} & \bd{0.132M} \\ 
		\hline
	\end{tabular}  
	
	\label{tab:speed_vgg} 
\end{table} 
\subsubsection{Main Results} 
We implement our algorithm on VGG-1-32 with R-FCN detector. We compare our method with  Li \etal \cite{li2017mimicking}, He \etal \cite{he2017channel} and group convolution based accelerating method including using Depthwise Convolution and Group Convolution. The results are shown in Table \ref{tab:vgg_other_method} (we set input as $1000\times600$ and compute the complexity). For fair comparison, we use the same implementation details for all experiments. We involve Depthwise Convolution and Group Convolution into VGG-1-32 structures, guaranteeing the similar complexity with the original network. For example, we extend the channel numbers of every convolution layers $c$ to $\lceil\sqrt{3}c\rceil$ and we set group number as 3. We also compare with pruning methods \cite{he2017channel} and \cite{molchanov2016pruning}. The pruning ratio is set as 8, which means the model we get after pruning has the same size with VGG-1-32.  
\begin{table}[tb]
	\tablestyle{4.0pt}{1.2}
	\newcommand{\tabincell}[2]{\begin{tabular}{@{}#1@{}}#2\end{tabular}}
	\centering
	\caption{ Comparison with other methods. The results show that our method outperforms others (higher is better). Group convolution based approach(Depthwise Convolution and Group Convolution) don`t work well on the very tiny model. Quantization Mimic also outperforms than Li \etal , who only uses mimic learning.}
	\begin{tabular}{c|c|p{0.8cm}<{\centering}p{0.8cm}<{\centering}p{0.8cm}<{\centering}}
		\hline
		solution & \tabincell{c}{complexity\\(MFLOPS)}&easy & medium & hard  \\
		\hline
		\hline
		scratch & 227 &71.3 & 55.4 & 23.8\\
		\hline
		Depthwise Convolution & 232 & 69.1 & 51.1 & 21.6 \\
		\hline
		Group Convolution(group 2) & 286 & 67.8 & 51.9 & 22.4 \\
		\hline
		Group Convolution(group 3) & 273 & 65.8 & 50.8 & 22.1 \\
		\hline
		He \etal \cite{he2017channel} & 227 & 68.0 & 50.7 & 22.1 \\
		\hline
		Molchanov \etal \cite{molchanov2016pruning} & 227 & 73.2 &58.2 & 25.2\\
		\hline
		Li \etal \cite{li2017mimicking}(only mimic) & 227 & 71.9 &58.2 & 25.6\\
		\hline
		Quantization Mimic & 227 & \bd{73.9} & \bd{62.1} & \bd{27.6}  \\
		\hline
	\end{tabular}  
	
	\label{tab:vgg_other_method}
\end{table} 
The results demonstrate that our algorithm outperforms other methods. We find that group convolution based methods are not suitable for very tiny networks. This is mainly because very tiny networks usually have small channel numbers and using group convolutions will block the information flow.  Compared with pruning methods \cite{molchanov2016pruning}\cite{he2017channel}, Quantization Mimic also works better. We argue that pruning methods can get good results on large models (\eg, VGG and Resnet). However, none of these works
try to prune a network to $\frac{1}{16}$ times.  Compared with mimic method \cite{li2017mimicking}, Quantization Mimic outperforms it by 2.0 points, 3.9 points and 1.9 points on \emph{easy}, \emph{medium} and \emph{hard} subsets. We find that quantized teacher network has better performance than full-precision teacher network. Ablation experiments are conducted to diagnose how Quantization Mimic brings improvement.

Table \ref{tab:vgg_self} further shows the effectiveness of our approach. We can see that our method can increase AP of very tiny models by 2.6 points, 6.7 points and 3.7 points on \emph{easy}, \emph{medium} and \emph{hard} subsets respectively. Results on \emph{medium} and \emph{hard} subsets, the small model can even achieve comparable results with large model.
\begin{table}[tb]
	\tablestyle{4.0pt}{1.2}
	\centering
	\caption{ The comparision between VGG and VGG-1-4 on the WIDER FACE dataset. We suggest that VGG has abundant structures and it has similar performance with VGG-1-4. And we choose VGG-1-4 as teacher model. }
	\begin{tabular}{c|c|p{0.8cm}<{\centering}p{0.8cm}<{\centering}p{0.8cm}<{\centering}}
		\hline
		Model & solution & easy & medium & hard \\
		\hline
		\hline
		VGG & full-precision & 83.9 & 61.0 & 26.8\\
		\hline
		\multirow{2}*{VGG-1-4} & full-precision & 82.4 & 62.5 & 26.3 \\
		\cline{2-5}
		~ & quantized & 83.7 & 65.0 & 27.4 \\
		\hline
		\multirow{2}*{VGG-1-32} & scratch & 71.3 & 55.4 & 23.8 \\
		\cline{2-5}
		~ & Quantization Mimic & 73.9 & 62.1 & 27.6 \\
		\hline
	\end{tabular}  
	
	\label{tab:vgg_self} 
\end{table} 

\subsubsection{Ablation Study on Quantization Operation} \label{ablation_quantization}
To verify the effectiveness of quantization operation, we do several experiments. As demonstrated in Table \ref{tab:vgg_ablation_quantization}, the performance of teacher network directly impact the performance of student network. Also, the quantization operation help mimic learning and improves the performance of student network. For the same quantized teacher network, doing quantization operation on the student network increase AP by 0.9 point, 2.8 points and 2.0 points on three subsets.
\begin{table}[tb]
	\tablestyle{4.0pt}{1.2}
	\newcommand{\tabincell}[2]{\begin{tabular}{@{}#1@{}}#2\end{tabular}}
	\centering
	\caption{Quantization \vs Nonquantization: The ablation study shows that the performance of student network depends on the performance of teacher network. The results also suggest that quantization method do help mimic learning.}
	\begin{tabular}{c|c|p{0.8cm}<{\centering}p{0.8cm}<{\centering}p{0.8cm}<{\centering}}
		\hline
		\tabincell{c}{ teacher  \\ quantization?}& \tabincell{c}{ student  \\ quantization?} & easy & medium & hard  \\
		\hline
		\hline
		& &71.9 & 58.2 & 25.6 \\
		\hline
		\checkmark &  & 73.0 & 59.3 &25.6 \\
		\hline
		\checkmark &\checkmark & \bd{73.9} & \bd{62.1} & \bd{27.6} \\
		\hline
	\end{tabular}  
	
	\label{tab:vgg_ablation_quantization} 
\end{table} 

We notice that quantization operation has regularization effect on network. To exclude that it is the regularization that bring improvement of performance, we also do experiments with and without quantization on student network. In Table \ref{tab:vgg_ablation_qua}, we find that only doing quantization has no influence on the performance, \ie , the improvement comes from Quantization Mimic.
\begin{table}[h]
	\tablestyle{4.0pt}{1.2}
	\centering
	\caption{ The influence of quantization only on small networks. The results suggest quantization only does not bring improvement.}
	\begin{tabular}{c|c|p{0.8cm}<{\centering}p{0.8cm}<{\centering}p{0.8cm}<{\centering}}
		\hline
		Model & quantization? & easy & medium & hard \\
		\hline
		\hline
		\multirow{2}*{VGG-1-32}& \checkmark & 71.9 & 55.2 & 23.7 \\
		\cline{2-5}
		~ &  &  71.3 & 55.4 & 23.8 \\
		\hline
	\end{tabular}  
	
	\label{tab:vgg_ablation_qua} 
\end{table} 

\begin{figure}[tb]
	\centering
	\setlength{\belowcaptionskip}{-10pt}
	\includegraphics[ width=0.48\linewidth]{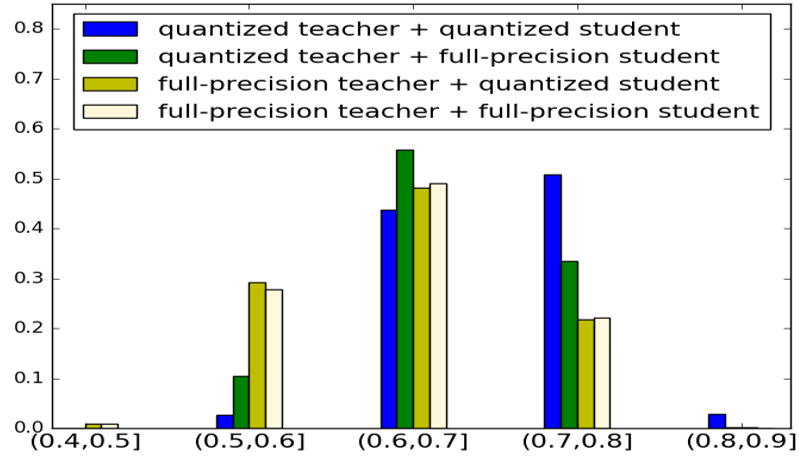}
	\caption{Histogram of matching ratio. The plot suggests that using quantiation operation both on teacher and student networks can help student network's feature maps to better match teacher network's. The horizontal axis represents bin of matching ratio, \ie the percentage of matched entries in a RoI. The vertical axis represents the frequency of RoIs within this bin. }
	\label{fig:matching}
\end{figure}

To further show that quantization operation can help student networks learn better, we illustrate the matching ratio of each RoI. In \S\ref{sec:analysis} we show that quantization operation promotes feature map matching between two networks. And in \S\ref{sec:mimic}, we introduce that our mimic learning is based on RoIs. Thus, we consider the matching ratio of each RoI, \ie, the percantage of elements in a RoI whose distance between two feature maps smaller than a threshold. We define the distance between $i$th entries of two feature maps as $|f_t^{i}-f_s^{i}|$, where $f_t^{i}$ and $f_s^{i}$ are the $ith$ element of teacher and student feature maps. If this distance is smaller than a threshold (we set 0.3 in this paper), then these two entries match. We evaluate on the validation set of WIDER FACE. We compare the results between full-precision and quantized network. The horizontal axis represents bin of matching ratio, i.e. the percentage of matched entries in a RoI. Figure \ref{fig:matching} demonstrates the results. The result shows that quantization operation can increase matching ratio of RoIs and promote feature maps matching process. Thus, quantization operation can help mimic learning.

\subsubsection{Ablation Study on Quantization Method}  
Different quantization method will bring different effects. The quantization methods we use in our work is uniform quantization. Another popular quantization method is power of 2 quantization, constraining the output to be either zero or power of 2. Table \ref{tab:vgg_ablation_uniform} illustrates the comparison of uniform quantization and power of 2 quantization. Teacher networks using different quantization methods have similar performance. However, the student network using uniform quantization is much better than using power of 2 quantization. We think this is probably because that our mimic learning is based on RoIs and strong responses in these areas are more important. So we should describe large number more accurately. And for power of 2 quantization method, it describes small numbers (\eg the number less than 1) accurately but roughly for large numbers. Thus, uniform quantization method is more reasonable and can bring better results. 
\begin{table}[tb]
	\tablestyle{4.0pt}{1.2}
	\centering
	\caption{Uniform Quantization \vs Power of 2 Quantization: Using uniform quantization as quantization method can get better result than using power of 2 quantization. }
	\begin{tabular}{c|c|p{0.8cm}<{\centering}p{0.8cm}<{\centering}p{0.8cm}<{\centering}}
		\hline
		Model & Quantization method & easy & medium & hard \\
		\hline
		\hline
		VGG-1-4 & power of 2 & 83.9 & 64.8 & 27.8 \\
		\cline{2-5}
		(teacher) & uniform(stride:1) & 83.7 & 65.0 & 27.4 \\
		\hline
		VGG-1-32 & power of 2 & 73.0 & 59.5 & 26.6 \\
		\cline{2-5}
		(student) & uniform(stride:1) & 73.9 & 62.1 & 27.6 \\
		\hline
	\end{tabular}  
	
	\label{tab:vgg_ablation_uniform} 
\end{table}

\subsection{Experiments on Pascal VOC Dataset}
We also carry out experiments on more complicated common object detection task. In this section we implement our approach on Resnet18 with Faster R-CNN detector for Pascal VOC object detection benchmark \cite{everingham2010pascal}. The experiments show that Quantization Mimic can extend to more complicated tasks.

Following \cite{ren2015faster}, we use Pascal VOC 2007 test set for test and trainval images in VOC 2007 and VOC 2012 for training (07+12). Hyperparameters in Faster R-CNN are same as \cite{ren2015faster}. Mean Average Precision (mAP) is used as the criterion to evaluate the performance of model. We use Resnet18 with Faster R-CNN framework as teacher networks. And Resnet18-1-16 with Faster R-CNN framework are selected as student networks accordingly.  We aim at improving the performance of the student works using Quantization Mimic method.

\subsubsection{Main Results}
First we compare the model using Quantization Mimic method with the model trained from scratch . Because of the poor learning ability of very tiny models, it is difficult to train them on complicated task, such as classification on Imagenet \cite{deng2009imagenet} and common object detections on Pascal VOC. Our method can improve a large margin of performance for very tiny networks on common object detections. Table \ref{tab:resnet_self} illustrates the results. We suggest that our method increase mAP  6.5 points for Resnet18-1-16 with Faster R-CNN framework. Relatively, we improve the performance for  $16.0\%$. The experiments also show that Quantization Mimic is easy to implement and can be extended to different frameworks. 
\begin{table}[tb]
	\centering
	\caption{ The comparision between Resnet18-1-16 with Faster R-CNN detector finetuned on Imagenet dataset and  using Quantization Mimic method. Our method can also bring huge improvement for very tiny networks on complicated common object tasks. }
	\begin{tabular}{c|c|c}
		\hline
		Model & solution & mAP \\
		\hline
		\hline
		\multirow{2}*{Resnet18} & full-precision & 72.9 \\
		\cline{2-3}
		~ & quantized & 73.3 \\
		\hline
		\multirow{2}*{Resnet18-1-16} & scratch & 40.5 \\
		\cline{2-3}
		~ & Quantization Mimic & 47.0  \\
		\hline
	\end{tabular}  
	
	\label{tab:resnet_self} 
\end{table}

We also do experiments compared with other accelerating and compressing methods. Same as the experiments on WIDER FACE dataset, we compare our method with Li \etal \cite{li2017mimicking}, who only use mimic learning. In Table \ref{tab:resnet_other_method}, our method outperforms than Li \etal \cite{li2017mimicking}. Our results are  2.4 points higher than our backbone, Li \etal \cite{li2017mimicking} on Resnet-1-16, which is a large margin. 

\begin{table}[tb]
	\newcommand{\tabincell}[2]{\begin{tabular}{@{}#1@{}}#2\end{tabular}}
	\centering
	\caption{ Comparison with backbone on Resnet18-1-16 with Faster R-CNN framework. Our method outperforms our backbone Li \etal \cite{li2017mimicking} methods for Resnet18 (higher is better). }
	\begin{tabular}{c|c|c}
		\hline
		Model & solution & mAP  \\
		\hline
		\hline
		\multirow{2}*{Resnet-1-16}  & Li \etal \cite{li2017mimicking}(only mimic)  &  44.6 \\
		\cline{2-3}
		~ & Quantization Mimic & \bd{47.0} \\
		\hline
	\end{tabular}  
	
	\label{tab:resnet_other_method} 
\end{table}

\subsubsection{Ablation Study on Mimic Loss Weight}
We propose that very tiny networks can be sensitive to loss weight in multi-loss task.  We do this experiment on Resnet18-1-16 to find a suitable mimic loss weight. In Table \ref{tab:resnet_ablation_loss}, we can see that the result of $\lambda=1$ is much better than the result of $\lambda=0.1$ and $\lambda=10$. We suggest that if mimic loss is too small (\eg $\lambda=0.1$) , the effectiveness of mimic learning will decline. However, if we set mimic loss weight too large (\eg $\lambda=10$), the very tiny network will mainly focus the gradient produced by mimic loss and ignore other gradients. And we call this phenomenon as `gradient focus' phenomenon.

\begin{table}[h]
	\setlength{\abovecaptionskip}{-10pt}
	\centering
	\caption{Mimic Loss Weight $\lambda$: The results show that very tiny networks are sensitive to the mimic loss weight. Either too large or too small loss weight will decrease the effectiveness of mimic learning. }
	\begin{tabular}{c|c|c}
		\hline
		Model & mimic loss weight & mAP \\
		\hline
		\hline
		\multirow{3}*{Resnet18-1-16}&10 & 44.1 \\
		\cline{2-3}
		~ & 1 &  \bd{47.0} \\
		\cline{2-3}
		~ & 0.1 & 43.0 \\
		\hline
	\end{tabular} 
	
	\label{tab:resnet_ablation_loss} 
\end{table}

\section{Conclusion}
In this paper, we propose Quantization Mimic to improve the performance of very tiny CNNs. We show quantization operation on both teacher and student networks can promote feature map matching. It becomes easier for the student network to learn after quantization operation. The experiments on WIDER FACE and Pascal VOC dataset demonstrate that quantization mimic outperforms state-of-the-art methods. We hope our approach can facilitate future research on training very tiny CNNs for cutting-edge applications.

\clearpage

\bibliographystyle{splncs}
\bibliography{2007}
\end{document}